# Azhary: An Arabic Lexical Ontology

Hossam Ishkewy, Hany Harb and Hassan Farahat

Azhar University, Faculty of Engineering, Computers and Systems Engineering Department

***Abstract***

*Arabic language is the most spoken languages in the Semitic languages group, and one of the most common languages in the world spoken by more than 422 million. It is also of paramount importance to Muslims, it is a sacred language of the Islamic Holly Book (Quran) and prayer (and other acts of worship) in Islam is performed only by mastering some of Arabic words. Arabic is also a major ritual language of a number of Christian churches in the Arab world and it is also used in writing several intellectual and religious Jewish books in the Middle Ages. Despite this, there is no semantic Arabic lexicon which researchers can depend on. In this paper we introduce Azhary as a lexical ontology for the Arabic language. It groups Arabic words into sets of synonyms called synsets, and records a number of relationships between words such as synonym, antonym, hypernym, hyponym, meronym, holonym and association relations. The ontology contains 26,195 words organized in 13,328 synsets. It has been developed and contrasted against AWN which is the most common available Arabic lexical ontology.*

***keywords:*** *Arabic Language, Ontology, WordNet, Semantic Web, Arabic Lexicon.*

## 1. Introduction
## 1.1 The Arabic Language

The language is a fundamental value in the life of every nation where it carries ideas and concepts and establishes the communication links among the people. The language templates which ideas, images and words that are formulated in which feelings and emotions never inseparable intellectual and emotional content[1].

The Arabic language is considered one of the most prolific languages in terms of language material, for example, The Lisan El-Arab (The Arabic Tongue) dictionary developed in the 13th century, contains more than 80,000 items, while in English, the dictionary of Samuel Johnson, which is considered one of the first English dictionaries, developed in the 18th century, contains of 42,000 words.

The Arabic language contains 28 written characters or 29 characters if the Hamza is considered as a separate character. The Arabic language is written from right to left like Persian, Hebrew, unlike many international languages and from the top to the bottom. It should be noted that the Arabic language includes Mandarin and Dialects, Mandarin is the language of the Quran, books and news bulletins and solemn occasion. Mandarin is the same in all Arab countries. The Arabic Dialects, is the language of daily communication, which vary from one country to another, for example, the Egyptian Dialect is different from the Iraqi Dialect. The most important differences between Mandarin and Dialect are that the first is written and spoken, while the second is only spoken, as well as that the norms of Mandarin rules, are fixed. This phenomena arises a debate around usage of Mandarin and Dialect between intellectuals where some support the use of Mandarin in all areas, and never use Dialect, while others prefer the use of Dialect instead of Mandarin.

Semantics plays an important role of Arabic language processing, we can't imagine accomplishing deep Arabic text processor, without sufficient information on the significance of the semantic relationships between words. The synonym, antonym, hypernym, hyponym, meronym, holonym and association relations in Arabic language are considered important linguistic features for the Arabic linguists. This research shows the importance of these relationships as a key to solving many of the linguistic analysis issues. The linguistic analysis issues are such as the development of characteristics and semantic fields of words, the text automatic analysis, the text understanding, solving the issues of linguistic confusion, and the machine translation from Arabic to another language or vice versa.

## 1.2 The Semantic Web

Current web contains billions of documents and has many administrative problems and limitations. Its content is still readable only by humans. To solve these problems, Tim Berners-Lee introduced the Semantic Web as a conceptual model of web that makes the contents to be read and used by human and intelligently by machines [2]. He mentioned that "*The Semantic Web (SW) is not a separate Web but an extension of the current one, in which information is given well defined meaning. Adding semantics to the Web involves allowing documents which have information in machine readable forms, and allowing links to be created with relationship values*"[3].

Tim Berners-Lee introduced in 2006 the fourth version of the Semantic Web Architecture containing eight layers [4]. The RDF and Ontology are considered to be the most important layers of this structure. The Resource Description Framework (RDF) is considered to be the key infrastructure to construct the Semantic Web , because RDF enables semantic interoperability. It is a standardized basis to encode , exchange and reuse the structured metadata . It is predicted that when RDF be used in a large scale on the web, content and relationship between different resources will be described better, and this will help search engines to easily find resources on the web and enable content to be rated [5]. Ontology is the backbone of the Semantic Web. ontology became an interesting topic for researches and an important key in the developing of the Semantic Web as it provides a sharable domain that facilitates the data understanding between people and different applications. Ontology has many definitions one of them defined by Gruber as "*an explicit specification of a conceptualization*" [6].

Since that *Tim Berners-Lee* invented the Semantic Web in 2001 and until now many researches about linguistic ontology have appeared, however still the linguistic ontology in Arabic language is fewer.

### 1.3 The Lexical Ontologies

WordNet is a lexical database for the English language. It collects English words into sets of synonyms called synsets, gives short definitions and usage examples, and records a number of relations among these synonym sets. These relations include: hypernym, hyponym, coordinate terms, meronym, holonym[7]. WordNet can be interpreted and used as a lexical ontology. It has been used for a number of different purposes in information systems, including word sense disambiguation, information retrieval, automatic text classification, automatic text summarization, machine translation and even automatic crossword puzzle generation[8]. The WordNet success of the English language stimulated similar projects that aim to develop WordNets for other languages, so, many WordNets has been carried out as an important resource for a wide range of natural language processing applications[9]. The global Wordnet (Global WordNet Organization), as non-profitable public organization, is established to support and encourage the development of dictionaries and WordNet for other languages based on the English WordNet, providing connectivity between the WordNet dictionaries from different languages such as Arabic, Hebrew, Persian, African, Albanian, Indian, etc.[10].

This research presents an Arabic WordNet (AWN) which is based on the design and contents of the universally accepted WordNet and mappable straightforwardly onto PWN 2.0 and EuroWordNet (EWN), enabling translation on the lexical level to English and dozens of other languages. The AWN was developed and linked with the Suggested Upper Merged Ontology (SUMO), where concepts are defined with machine interpretable semantics in first order logic [11].

This paper is arranged as it follows. Section 1 introduces the Arabic language, the Semantic Web and the Lexical Ontologies. Section 2 discusses some related works. Section 3 discusses our proposed Azhary system architecture and its implementation issues. Section 4 evaluates the system. The paper is concluded in Section 5.

## 2. Related works

Many researches concerned with the Arabic ontologies construction have been appeared in the last few years. These ontologies belong to variant domains and are constructed with different methods.

### 2.1. Islamic Domain Ontology

Al-Yahya, and others [12] proposed a computational model for representing Arabic lexicons using ontologies. The ontology development based on the UPON (Unified Process for ONtology) ontological engineering approach [13]. The ontology was limited to time nouns which appeared in the Holy Quran. The ontology consisted of 18 classes and contained a total of 59 words. Hikmat Ullah Khan and others [14] proposed that the ontology concept can be applied for developing semantic search in Holy Quran. They presented a domain ontology, based on living creatures including animals and birds mentioned in Holy Quran. Authors in [15] explored the representation and classification of Holy Quran knowledge by using ontology. The ontology model for Al-Quran was developed according to Al-Quran knowledge themes as described in Syammil Al-Quran Miracle the Reference. Iman (faith) and Akhlaq (deed) main classes were chosen as the research scope for constructing the ontology.

## 2.2. Linguistic Ontology

Authors in [16] described the Semantic Quran dataset as a multilingual RDF representation of translations of the Quran. The dataset was created using two different semi-structured sources(The Tanzil Project, The Quranic Arabic Corpus Project). The dataset was aligned to an ontology designed to represent multilingual data from sources with a hierarchical structure. The resulting RDF data encompassed 43 different languages which are belonged to the most under represented languages in Linked Data, including Arabic, Amharic and Amazigh. The authors presented the ontology devised for structuring the data. It included four basic classes: Chapter, Verse, Word and LexicalItem. Belkredim and El Sebai [17] developed Arabic ontology using verbs and roots, and classified verbs of derivation in the Arabic language from the roots based on that 85% of words derived from tri-literal roots. The use of the roots as a base for building ontology is inaccurate because the word derived, though they have the same basic meaning, but may cannot be grouped under the same categories. Also, the authors did not provide any implementation. Author in [18] aimed to develop an Arabic Linguistic or Upper Ontology. The top levels of the Arabic ontology was built manually based on DOLCE and SUMO upper level ontologies. Only 420 concepts of the Arabic ontology is being evaluated, and the remainder concepts did not finished. Authors in [19] presented Al –Khalil project for building an Arabic infrastructure ontology. The core of this infrastructure is a linguistic ontology that is founded on Arabic traditional grammar. The development of Al –Khalil contained two steps. The first step is bootstrapping manually the ontology by choosing the linguistic concepts from Arabic linguistics and relating them to the concepts in GOLD. The second is using an automatic extraction algorithm to extract new concepts from linguistic texts to enrich the ontology.

After the great success of the WordNet, many WordNets appeared in several languages until Arabic WordNet (AWN) developed following a methodology developed for EuroWordNet. Arabic WordNet consists of 11,270 synsets and contains 23,496 Arabic expressions (words and multiwords) [20]. Many researchers depended on AWN in their work such as [21] and [22] in Q/A system. Some researchers tried to improve and extend AWN. Authors in [23] extended the AWN using lexical and morphological rules and applying Bayesian inference. Alkhalifa and Rodríguez [20] used Wikipedia to automatically extending named entities of Arabic WordNet while in [24] authors used the Yago ontology as a resource for the enrichment of named entities in Arabic WordNet.

## 2.3. Automatic Ontology Generation

[25] presented an approach to the automatic generation of ontology instances from a collection of unstructured known documents as Al-Quran. The presented approach was stimulated based on the combination of Natural Language Processing techniques, Information Extraction and Text Mining techniques. [26] presented an approach to automatic ontology construction from a corpus of domain "Arabic linguistics". They reused information extraction techniques for extracting new terms that will denote elements of the ontology(concept, relation). [27] illustrated an ontology extraction based on fuzzy-swarm algorithm for Islamic knowledge text. The authors supposed that combining lexicon, syntactic and statistical learning methods, the accuracy and the

computational efficiency of the ontology discovery process is improved. [28] proposed a system that automates the process of constructing a taxonomic agricultural domain ontology using a semi-structured domain specific web documents.

### 2.4. Miscellanies Ontologies

[29] presented multilingual tool providing online Arabic information retrieval based on legal domain ontology under try to improve the search recall and precision . They use the built ontology as a base for subsequent user-query expansion in the search system. In [30] the authors provided an Arabic ontology representing computer technology knowledge based on the Arabic Internet blogs. The research is conducting a pilot study on a number of Arabic blogs randomly selected in computer technology. The ontology comprised 110 classes, 78 class instances and 48 object properties while others [31] also developed an ontology in the computer technology domain. This ontology was built based on classic Arabic language rather than the blogs modern language. The presented ontology is much simpler than what is discussed in [30].

## 3. The Azhary System Architecture

**Azhary** is a lexical ontology for the Arabic language, its primary use is in automatic text analysis and artificial intelligence applications. It groups Arabic words into sets of synonyms called *synsets*, and records a number of relations between words. The ontology contains 26,195 words, organized in 13,328 *synsets*. The Azhary system architecture as shown in Figure 1 is composed of different modules: The Words Extraction, The Relations Building, The Ontology Building, The Searching, and The Graphical User Interface.

### 3.1 The Words Extraction

The words extraction is the process of creating the seed words to start the ontology. This seed is built from the Holy Quran which has 77,439 words. The primary function of this process is to cut the Quran words and storing these words with no repetition into an excel file which contains the words and the relations in tabular form. The Quran text is extracted from Tanzil [32] (an international Quranic project) which provides a highly-verified precise Quran text.

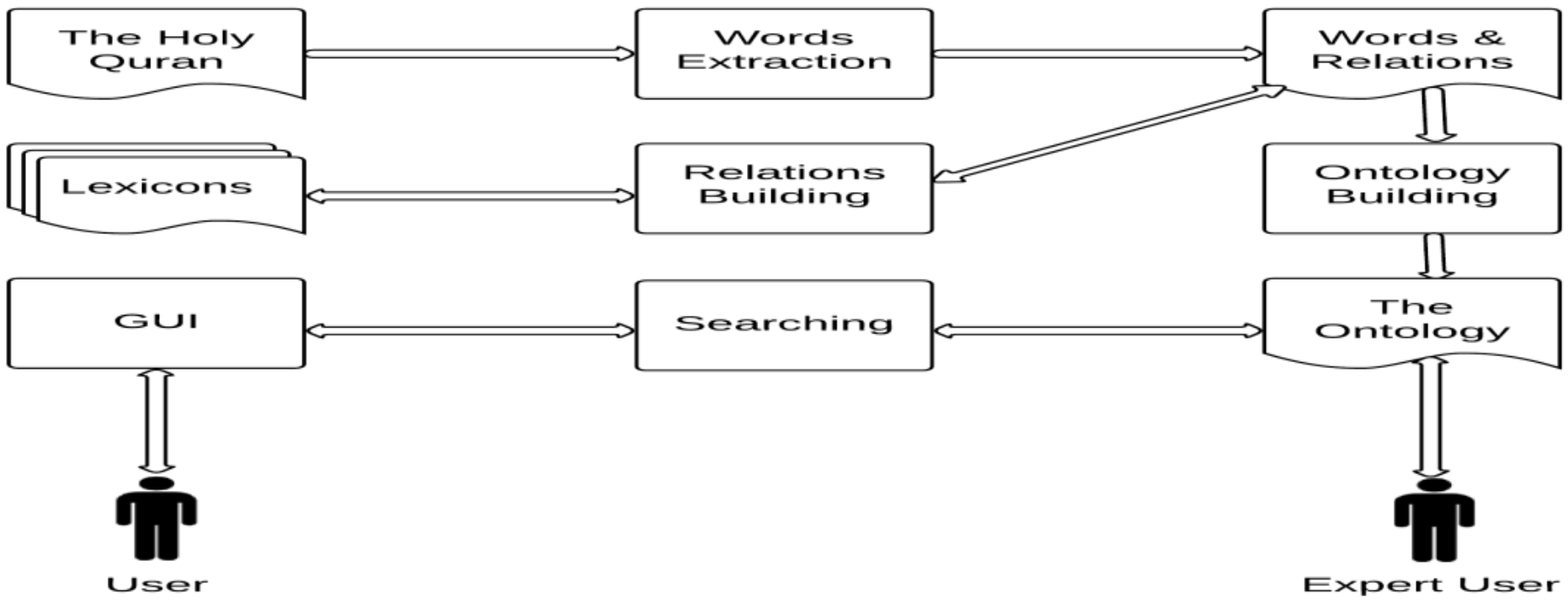

Figure 1 :The Azhary System Architecture

## 3.2 The Relations Building

The relations building module is a fully manual process where it is achieved by using Arabic/Arabic dictionaries to find the words and build the relations between them. Common dictionaries, such as the meanings dictionary (قاموس المعانى) [33] and rich lexicon(معجم الغني) [34] and mediator lexicon (المعجم الوسيط) [35] are used in the construction of the ontology. The word can be connected to another word by means of the following semantic relations :

- **synonym:** B is a synonym of A, if A and B has the same meaning.
  Ex: (مَعْرُوف ) is a synonym of (جَمِيل). Both mean kindness
- **hypernym:** B is a hypernym of A, if A is a (kind of) B.
  Ex: limb(عُضو) is a hypernym of hand (يَد)
- **hyponym:** B is a hyponym of A, if B is a (kind of) A .
  Ex: hand (يَد) is a hyponym of limb (عُضو)
- **meronym:** B is a meronym of A, if B is a (part of) A .
  Ex: hand (يَد) is a meronym of arm(ذِراع)
- **holonym:** B is a holonym of A, if A is a (part of ) B .
  Ex: arm(ذِراع) is a holonym of hand (يَد)
- **antonym:** B is an antonym of A, if A is an (inverse) of B.
  Ex: kindness (مَعْرُوف ) is an antonym of harm (أذيَة)
- **association:** A and B are associated if A exists always with B.
  Ex: doctor (طبيب ) and patient (مريض) are associated. husband(زوج) and wife(زوجة) are also associated.

**Azhary** includes parts of speech (POS) of the Arabic language defined by [36]. Figure 2 represents these POS illustrated by the free open source ontology editor protégé [37]. Figure 3 translates this POS to English language.

### 3.3 The Ontology Building.

The ontology building module reads the words and their relations from the excel file using JExcelApi [39]. JExcelApi is an open source java API enabling developers to read, write, and modify Excel spreadsheets dynamically. This process converts the words and relations tabular form to an ontology. Azhary ontology has been developed using the free and open source java framework for semantic web building (Apache Jena) [39]. A simple example of the ontology building process is listed below to explain how the ontology has been built.

```
<rdf:Description rdf:about="http://www.azhary.org#يَد">
 <rdf:type rdf:resource="http://www.azhary.org#اسم"/>
   <a:anti rdf:resource="http://www.azhary.org#ضَرَرٌ"/>
   <a:anti rdf:resource="http://www.azhary.org#شَرٌّ"/>
```

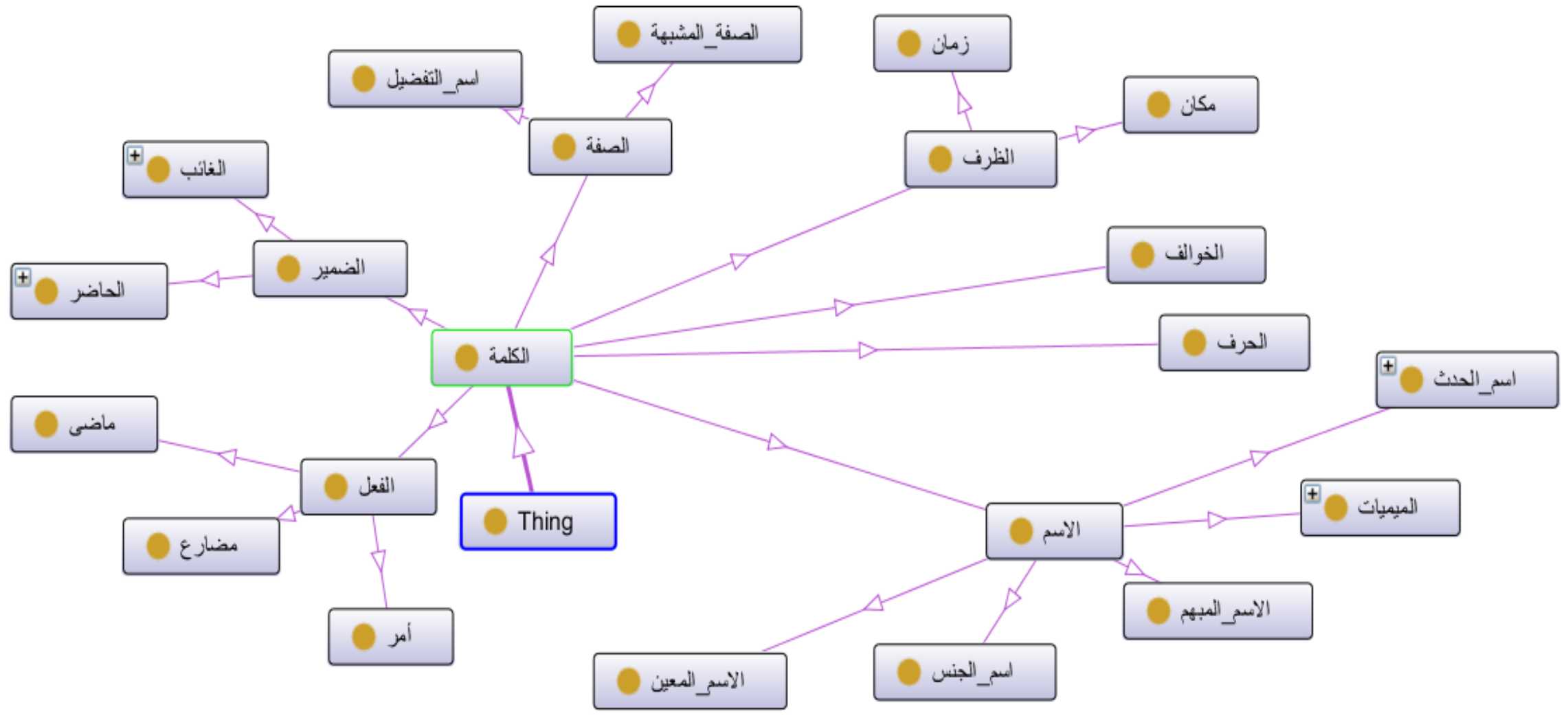

Figure 2: The Arabic POS

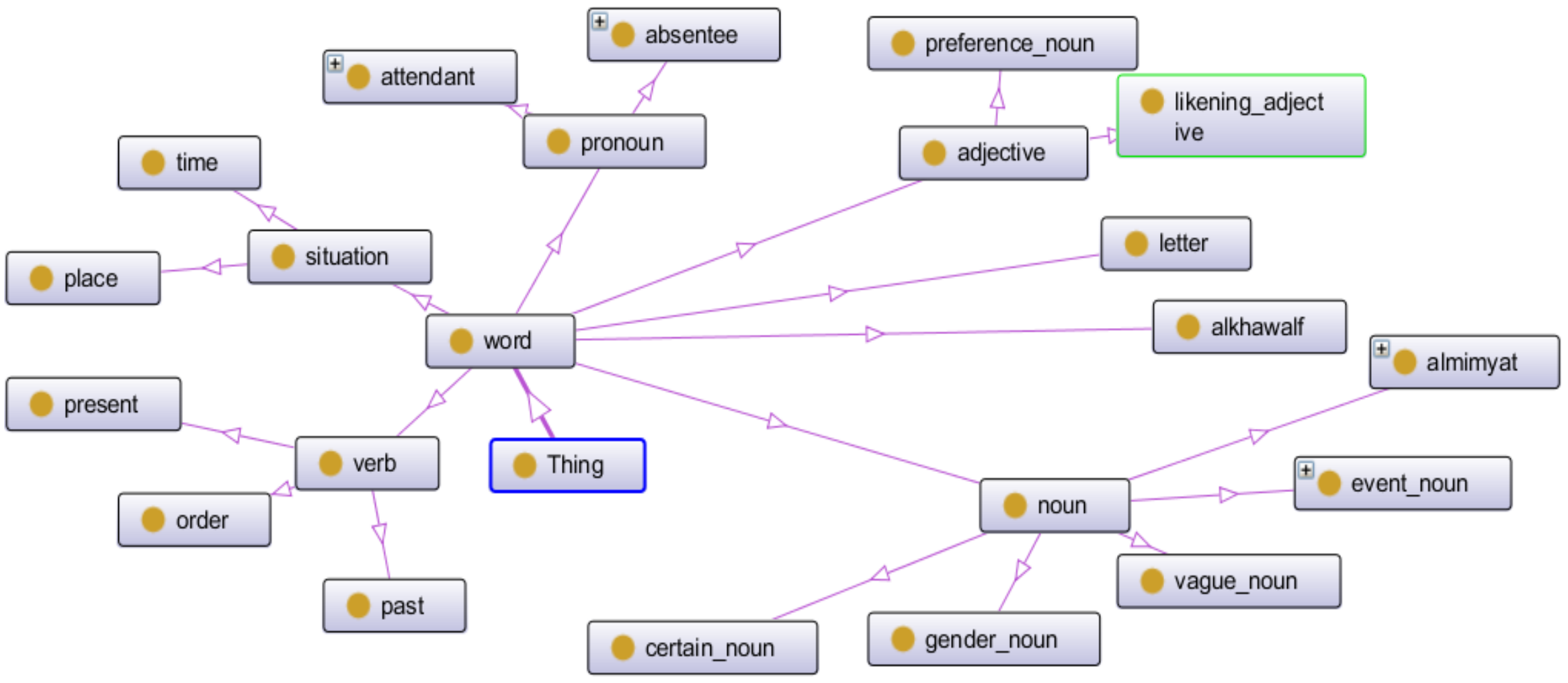

Figure 3: Arabic POS translated into English

<a:means rdf:resource="http://www.azhary.org#سُلطانٌ"/>
<a:has_a rdf:resource="http://www.azhary.org#بُنصُر"/>
<a:part_of rdf:resource="http://www.azhary.org#ذِراع"/>
<a:part_of rdf:resource="http://www.azhary.org#جِسم"/>
<a:has_a rdf:resource="http://www.azhary.org#خُنصُر"/>
<a:means rdf:resource="http://www.azhary.org#إِحْسان"/>
<a:has_parent rdf:resource="http://www.azhary.org#عُضو"/>
<a:means rdf:resource="http://www.azhary.org#صَدَقَة"/>
<a:means rdf:resource="http://www.azhary.org#مِلْك"/>
<a:has_a rdf:resource="http://www.azhary.org#سَبَابَة"/>
<a:anti rdf:resource="http://www.azhary.org#سُوءٌ"/>

```
  <a:has_child rdf:resource="http://www.azhary.org#يَد_يُسرَى"/>
  <a:has_a rdf:resource="http://www.azhary.org#سُلاميات"/>
  <a:has_a rdf:resource="http://www.azhary.org#إبهام"/>
  <a:means rdf:resource="http://www.azhary.org#قُوَّة"/>
  <a:has_parent rdf:resource="http://www.azhary.org#أداة"/>
  <a:has_a rdf:resource="http://www.azhary.org#رُسغ"/>
  <a:anti rdf:resource="http://www.azhary.org#أَذِيَّةٌ"/>
  <a:means rdf:resource="http://www.azhary.org#مَعْرُوف"/>
  <a:means rdf:resource="http://www.azhary.org#جَمِيل"/>
  <a:has_a rdf:resource="http://www.azhary.org#كَف"/>
  <a:has_a rdf:resource="http://www.azhary.org#وُسطَى"/>
  <a:has_child rdf:resource="http://www.azhary.org#يَد_يُمنَى"/>
  <a:has_parent rdf:resource="http://www.azhary.org#طَرف"/>
  <a:means rdf:resource="http://www.azhary.org#مَقْبَضُ"/>
  <a:means rdf:resource="http://www.azhary.org#قُدرة"/>
  <a:means rdf:resource="http://www.azhary.org#نِعْمَة"/>
</rdf:Description>
```

The previous ontology part describes the resource hand(يَد), its part of speech is noun(اسم). The synonym relation is represented by the means property. The hypernym relation is represented by the has_parent property. The hyponym relation is represented by the has_child property. The meronym relation is represented by the part_of property. The holonym relation is represented by the has_a property. The antonym relation is represented by the anti property. Figure 4 introduces Azhary GUI of the example. Azhary GUI consists of sixteen components ( two text fields, six text area and eight labels describing the purpose of the text areas and fields). Semantic experts can query this ontology easily but normal user can only use the GUI.

## 3.4 The Searching Process

The searching process goal is to receive the use query from the GUI, query the ontology to find the words and relations, return the results to the GUI. We used (Apache Jena) [39] also in this querying.

## 3.5 The Graphical User Interface.

The GUI is used only to receive the user query and return the results. Figure 4 shows the Azhary GUI and Table 1 translates the GUI labels from Arabic to English.

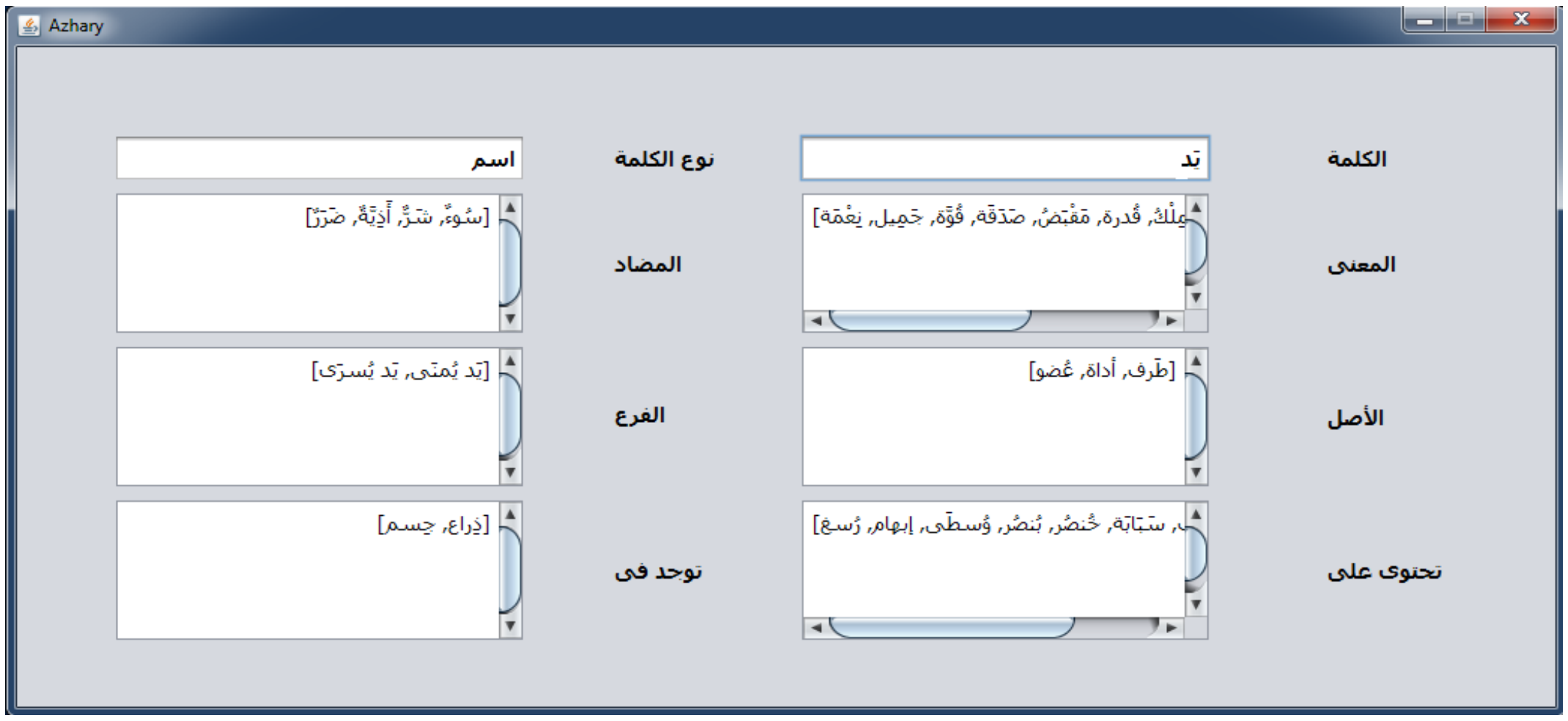

Figure 4: The Azhary GUI

| الكلمة | نوع الكلمة | المعنى | المضاد | الأصل | الفرع | تحتوى على | توجد فى |
|---|---|---|---|---|---|---|---|
| word | POS | synonym | antonym | hypernym | hyponym | holonym | meronym |

Table 1

## 4. Criticism and Comparison

AWN as an Arabic lexicon has the largest number of words and synsets in comparison with the other Arabic lexicons. But AWN has many disadvantage such that:

1- It has many errors in the meanings.

Ex: if you use AWN to find the meaning of the verb fail(فشِل) you will find (أصاب بِعَرَج) in the results. This is wrong Arabic sentence. The true sentence is (أُصِيب بِعَرَج) which means (be lame) and it does not mean fail.

2-It has many words repetition.

EX: if you use AWN to find the meaning of the verb speak(تَكَلَّمَ) or the noun traveling. (سَفَر), you will find many words repetition in the results.

3- A word may has different individual properties with itself.

EX: if you use AWN to find the meaning of the noun food (طَعَام) you will find that (طَعَام بخلاف طَعَام ) in the results which means (food is different from food).

4-AWN has errors in the Arabic word types tree.

Ex: The verb is considered as a subclass of the noun.

5-AWN ignores the letters which has important meaning in Arabic language.

EX: the letter(ليت) means (hopefulness).

6- Many sentences have stuck words.

EX:(منتجاتألبان) is two stuck words while the true sentence is (منتجات ألبان) means milk products.

7-AWN has some Arabic diacritics problems.

8-AWN is limited in terms of semantic meanings and the relations between words [22].

Table 2 shows a comparison between AWN and Azhary

| **Criteria\Lexicon** | **AWN** | **Azhary** |
|---|---|---|
| **Average Response Time** | 1.3 Second | 11 Millie Second |
| **Number of Words** | 23,496 | 26,195 |
| **Number of Synsets** | 11,270 | 13,328 |
| **Synonym Relation** | Yes | Yes |
| **Hypernym Relation** | Yes | Yes |
| **Hyponym Relation** | Yes | Yes |
| **Meronym Relation** | No | Yes |
| **Holonym Relation** | No | Yes |
| **Antonym Relation** | No | Yes |
| **Association Relation** | No | Yes |

# 5. The Conclusions

There is no chance for Arabic semantic analysis since there is no Arabic lexical ontology can linguist researchers depend on, So, we have introduced in this paper Azhary as a lexical ontology for the Arabic language, its primary use is in automatic text analysis and artificial intelligence applications. It groups Arabic words into sets of synonyms called *synsets*, and records some relations between words such as: the synonym, antonym, hypernym, hyponym, meronym, holonym and association relations. It has been developed and contrasted against AWN which is the most common available Arabic lexical ontology. Our system shows better response time and it is has larger words and records more word relationships.

# 5. References

[1] Farhan El-Salim,"اللغة العربية ومكانتها بين اللغات ", Available: http://www.saaid.net/Minute/33.htm, [Accessed: Aug. 5, 2014].

[2] Haytham T. Al-Feel, Magdy Koutb and Hoda Suoror, "Semantic Web on Scope: A New Architectural Model for the Semantic Web",Journal of Computer Science 4(7): 613-624, 2008.

[3] Tim Berners-Lee, James Hendler and Ora Lassila,"The Semantic Web", Scientific American, May 17, 2001

[4] T. Berners-Lee, "Artificial Intelligence & the Semantic Web" World Wide Web Consortium, 2006. [Online]. Available: http://www.w3.org/2006/Talks/0718-aaai-tbl/Overview.html#(14) [Accessed: Sep. 5, 2014].

[5] Liyang Yu, "Introduction to the Semantic Web and Semantic Web Services",2007.

[6] T. R. Gruber, "A Translation Approach to a Portable Ontology Specification", Knowledge Acquisition, vol. 6, pp. 199-221, 1993.

[7] G. A. Miller, R. Beckwith, C. Fellbaum, D. Gross, and K. J. Miller " Introduction to WordNet: an on-line lexical database ", International Journal of Lexicography 3(4):235-244 (1990).

[8] http://en.wikipedia.org/wiki/WordNet [Accessed: Sep. 9, 2014].

[9] Aya M. Al-Zoghbya , Ahmed Sharaf Eldin Ahmed, Taher T. Hamza, "Arabic Semantic Web Applications – A Survey", Semantic Web Journal, 2012.

[10] http://globalwordnet.org/wordnets-in-the-world[Accessed: Sep. 7, 2014].

[11] Christiane Fellbaum, Musa Alkhalifa, William J.Black, Sabri Elkateb, Adam Pease, HoracioRodríguez, and Piek Vossen, "Building a WordNet for Arabic", May, 2006.

[12] Maha Al-Yahya, Hend Al-Khalifa,Alia Bahanshal, Iman Al-Odah, Nawal Al Helwah, "AN ONTOLOGICAL MODEL FOR REPRESENTING SEMANTIC LEXICONS: AN APPLICATION ON TIME NOUNS IN THE HOLY QURAN", The Arabian Journal for Science and Engineering, Volume 35, December 2010.

[13] A. De Nicola, M. Missikoff, and R. Navigli, "A Software Engineering Approach to Ontology Building",Information Systems, 34(2009), pp. 258–275.

[14] Hikmat Ullah Khan, Syed Muhammad Saqlain, Muhammad Shoaib,Muhammad Sher, "Ontology Based Semantic Search in Holy Quran", International Journal of Future Computer and Communication, Vol. 2, No. 6, December, 2013.

[15] Azman Ta'a, Syuhada Zainal Abidin, Mohd Syazwan Abdullah, Abdul Bashah B Mat Ali, and Muhammad Ahmad," AL-QURAN THEMES CLASSIFICATION USING ONTOLOGY", Proceedings of the 4the International Conference on Computing and Informatics, ICOCI, 2013.

[16] Mohamed Ahmed Sherif ,Axel-Cyrille Ngonga Ngomo, "Semantic Quran: A Multilingual Resource for Natural-Language Processing", Semantic Web Journal,2009

[17] Fatma Zohra Belkredim, Ali El Sebai "An Ontology Based Formalism for the Arabic Language Using Verbs and their Derivatives", IBIMA,Volume 11, 2009.

[18] Mustafa Jarrar, "Building a Formal Arabic Ontology" , In proceedings of the Experts Meeting on Arabic Ontologies and Semantic Networks April 26-28, 2011.

[19] Hassina Aliane, Zaia Alimazighi, Mazari Ahmed Cherif,"Al–Khalil: The Arabic Linguistic Ontology Project", Seventh International Conference on Language Resources and Evaluation, 2010.

[20] Musa Alkhalifa, Horacio Rodríguez, "Automatically Extending Named Entities Coverage of Arabic WordNet using Wikipedia ",International Journal on Information and Communication Technologies, Vol. 3, No. 3, June 2010.

[21] Lahsen Abouenour," On the Improvement of Passage Retrieval in Arabic Question/Answering (Q/A) Systems", Natural Language Processing and Information Systems, Volume 6716, 2011, pp 336-341.

[22] Abdurrahman A. Nasr., M.Sc. thesis, Al-Azhar university, Faculty Of Engineering, Computers and Systems Dept., 2012.

[23] Rodríguez, H. Farwell, D. Farreres, J. Bertran, M. Alkhalifa, M. Martí M.A " Arabic WordNet: Semi-automatic Extensions using Bayesian Inference". In Proceedings of the the 6th Conference on Language Resources and Evaluation LREC,May, 2008.

[24] Lahsen Abouenour, Karim Bouzoubaa, Paolo Rosso," Using the Yago ontology as a resource for the enrichment of Named Entities in Arabic WordNet ", Workshop on LR & HLT for Semitic Languages, 7th Int. Conf. on Language Resources and Evaluation, LREC, May, 2010.

[25] Saidah Saad, Naomie Salim and Hakim Zainal,"Islamic Knowledge Ontology Creation", Internet Technology and Secured Transactions, 2009.

[26] Ahmed Cherif Mazari , Hassina Aliane, and Zaia Alimazighi ,"Automatic construction of ontology from Arabic texts" ,Proceedings ICWIT, 2012.

[27] Saidah Saad, Naomi Salim, "Methodology of Ontology Extraction for Islamic Knowledge Text", In Postgraduate Annual Research Seminar, UTM,2008.

[28] Samhaa R. El-Beltagy, Maryam Hazman, Ahmed Rafea, "Ontology learning from domain specific web documents," vol. 4, No. 1/2, May 2009.

[29] S. Zaidi, M.T. Laskri, K. Bechkoum, "A Cross-language Information Retrieval Based on an Arabic Ontology in the Legal Domain, IEEE SITIS" ,2005.

[30] Lilac Al-Safadi, Mai Al-Badrani, Meshael Al-Junidey, "Developing Ontology for Arabic Blogs Retrieval", International Journal of Computer Applications (0975– 8887),Volume 19– No.4, April, 2011.

[31] Ibrahim Fathy Moawad, Mohammad Abdeen, Mostafa Mahmoud Aref, "Ontology-based Architecture for an Arabic Semantic Search Engine", December 15-16, 2010.

[32] Tanzil, http://www.tanzil.net

[33] The meanings dictionary(قاموس المعانى),http://www.almaany.com/

[34] Abd El-Ghany Abo El-Azm, rich lexicon(معجم الغني),2013

[35] Mediator Lexicon (المعجم الوسيط),http://kamoos.reefnet.gov.sy/
[36] Fadel Mostafa El-Saki, "Arab parts of speech in terms of form and function" (أقسام الكلام العربى من حيث الشكل والوظيفة),1977.
[37] Protege http://protege.stanford.edu/
[38] JExcelApi, http://jexcelapi.sourceforge.net/
[39] Apache Jena, https://jena.apache.org/index.html